**Fubini–Study geometry of representation drift in high-dimensional data**

Arturo Tozzi (corresponding author)
ASL Napoli 1 Centro, Distretto 27, Naples, Italy
Via Comunale del Principe 13/a 80145
tozziarturo@libero.it

ABSTRACT

High-dimensional representation drift is commonly quantified using Euclidean or cosine distances, which presuppose fixed coordinates when comparing representations across time, training or preprocessing stages. While effective in many settings, these measures entangle intrinsic changes in the data with variations induced by arbitrary parametrizations. We introduce a projective geometric view of representation drift grounded in the Fubini–Study metric, which identifies representations that differ only by gauge transformations such as global rescalings or sign flips. Applying this framework to empirical high-dimensional datasets, we explicitly construct representation trajectories and track their evolution through cumulative geometric drift. Comparing Euclidean, cosine and Fubini–Study distances along these trajectories reveals that conventional metrics systematically overestimate change whenever representations carry genuine projective ambiguity. By contrast, the Fubini–Study metric isolates intrinsic evolution by remaining invariant under gauge-induced fluctuations. We further show that the difference between cosine and Fubini–Study drift defines a computable, monotone quantity that directly captures representation churn attributable to gauge freedom. This separation provides a diagnostic for distinguishing meaningful structural evolution from parametrization artifacts, without introducing model-specific assumptions. Overall, we establish a geometric criterion for assessing representation stability in high-dimensional systems and clarify the limits of angular distances. Embedding representation dynamics in projective space connects data analysis with established geometric programs and yields observables that are directly testable in empirical workflows.

INTRODUCTION

High-dimensional data analysis relies on geometric notions of similarity and change, with Euclidean and cosine distances remaining the dominant tools for quantifying representation drift across time, preprocessing stages or learning iterations (Tang et al. 2012; Canessa et al. 2020; Feng et al. 2022; Li et al. 2025). These measures embed strong assumptions about the meaningfulness of coordinates and parametrizations. They implicitly treat representations as points in a fixed vector space, even when the underlying objects are only defined up to normalization, sign or more general transformations. Apparent changes in representation space may reflect gauge-induced variability rather than intrinsic evolution of the data (Green et al. 2021; Hu and Szymczak 2023; Han et al. 2023; Rahnenführer et al. 2023). This limitation has been described across diverse domains, including dimensionality reduction, latent variable modeling and representation learning, where equivalent solutions can differ substantially in their numerical realization while encoding the same structural information (Lesort et al. 2018; Sinha et al. 2021; Zhang et al. 2022; Goteti et al. 2024; Vo et al. 2024; Chiossi et al. 2025). Existing approaches usually address the issue through ad hoc normalization or alignment procedures, rather than by redefining the geometry in which representation drift is measured. As a result, there remains no criterion to separate intrinsic change from parametrization-induced drift in a way that is invariant by construction, which leaves a conceptual gap between empirical practice and the equivalence relations governing high-dimensional representations. This gap motivates the development of a distance notion that respects the projective nature of many data representations and allows their evolution to be compared in a meaningful way.

We introduce a projective geometric perspective on representation drift based on the Fubini–Study metric, which provides a natural distance on spaces of equivalence classes rather than on raw vectors (Elvang 2021; Espinosa-Champo and Naumis 2023; Hofmann et al. 2023). By treating representations as rays instead of points and measuring change solely in terms of intrinsic orientation, this metric explicitly identifies states differing only by gauge transformations. We operationalize this idea by tracking cumulative drift along empirical representation trajectories and comparing Euclidean, cosine and Fubini–Study distances computed along the same paths. This yields a computable quantity that quantifies the contribution of gauge-induced variability to observed drift, without requiring alignment, reference choices or model-dependent corrections. We designed experiments to test this distinction directly, using empirical datasets in which genuine projective ambiguity is present. Further, we assessed whether the resulting separation between angular and projective measures is systematic and robust. Taken together, these considerations establish the conceptual basis for a geometry of representation drift able to distinguish intrinsic evolution from parametrization artifacts.

We will proceed as follows. We first describe the construction of representation trajectories and the computation of Euclidean, cosine and Fubini–Study cumulative drift. We then present empirical results on freely available datasets with projective ambiguity. Finally, we discuss the implications of these findings for interpreting representation drift in high-dimensional data.

METHODS

We describe the mathematical constructions, data manipulations and computational procedures to quantify representation drift under different geometric notions of distance. We emphasize reproducible steps, detailing how representation trajectories are built, how distances are computed along them and how projective ambiguity is handled analytically and numerically.

**Dataset description**. All analyses were performed on the handwritten digits dataset distributed with the scikit-learn library. This dataset comprises 1,797 real grayscale images of handwritten numerical digits acquired from human subjects, each image represented as an $8 \times 8$ pixel grid and flattened into a 64-dimensional real-valued feature vector. Pixel intensities correspond to measured grayscale levels. The dataset includes class labels identifying the digit depicted in each image; however, labels are not used in the construction of representation trajectories. No intrinsic temporal ordering is provided by the data. An ordering is imposed solely to define successive analysis windows, enabling the study of representation drift under controlled and reproducible conditions. The choice of this dataset reflects its public availability, well-documented provenance, fixed dimensionality and suitability for principal component analysis without additional preprocessing beyond centering.

**Construction of representation trajectories**. The next step consists of converting empirical data into ordered trajectories suitable for geometric analysis. Let $X = \{x_i\}_{i=1}^{N} \subset \mathbb{R}^D$ denote a collection of empirical observations, where each $x_i$ is a $D$-dimensional feature vector. To study representation drift, the data are embedded into ordered sequences reflecting a notion of progression. In our analysis ordering is induced by a sliding-window procedure over the dataset index. For a fixed window length $W$ and step size $s$, contiguous subsets $X_k = \{x_k, x_{k+1}, \ldots, x_{k+W-1}\}$ are extracted for $k = 1, 1+s, 1+2s, \ldots$. Each subset defines a local representation of the data and the sequence of subsets induces a discrete trajectory indexed by $k$. This construction does not assume temporal causality; it only imposes an ordering sufficient to define successive representation states. Each state is then mapped to a vector in $\mathbb{R}^D$ or to a derived object, depending on the analysis step. The trajectory length $T$ is determined by the number of windows, $T = \lfloor (N-W)/s \rfloor + 1$. All subsequent geometric quantities are computed along this discrete trajectory $\{r_k\}_{k=1}^{T}$, where $r_k$ denotes the representation extracted from window $X_k$. This ordering enables the definition of stepwise changes $r_{k+1} - r_k$ and cumulative measures of drift along the trajectory, which form the basis of our comparative analysis across distance metrics.

**Euclidean distance and cumulative drift**. For representations $r_k \in \mathbb{R}^D$, Euclidean distance is defined as

$$d_E(r_k, r_{k+1}) = \| r_{k+1} - r_k \|_2 = \sqrt{\sum_{j=1}^{D} (r_{k+1}^{(j)} - r_k^{(j)})^2}.$$

This distance treats representations as points in a fixed vector space and is sensitive to both direction and magnitude. To characterize drift along the trajectory, stepwise Euclidean distances are accumulated to form a cumulative drift function

$$\Delta_E(n) = \sum_{k=1}^{n-1} d_E(r_k, r_{k+1}), n = 2, \ldots, T.$$

This quantity grows monotonically by construction and reflects the total path length traversed in $\mathbb{R}^D$. No normalization is applied unless explicitly stated, so changes in global scale contribute directly to $\Delta_E$. Euclidean cumulative drift provides a reference against which scale-invariant and projective measures can be compared. In practice, Euclidean distances are computed using vectorized linear algebra routines, ensuring numerical stability through double-precision arithmetic. The Euclidean measure is not intended to capture equivalence relations among representations; instead, it serves to quantify the raw magnitude of change under the chosen parametrization.

**Cosine distance and angular drift**. We introduce here angular comparison as a scale-invariant alternative. To remove sensitivity to global rescaling, representations are normalized to unit norm. For any nonzero vector $r \in \mathbb{R}^D$, define $\hat{r} = r/\| r \|_2$. The cosine similarity between successive states is then

$$s_C(r_k, r_{k+1}) = \langle \hat{r}_k, \hat{r}_{k+1} \rangle,$$

where $\langle \cdot, \cdot \rangle$ denotes the standard inner product. The corresponding angular distance is

$$d_C(r_k, r_{k+1}) = \arccos(\langle \hat{r}_k, \hat{r}_{k+1} \rangle),$$

with values in $[0, \pi]$. This distance measures the angle between normalized representations and is invariant under positive scalar multiplication. As with the Euclidean case, cumulative cosine drift is defined by

$$\Delta_C(n) = \sum_{k=1}^{n-1} d_C(r_k, r_{k+1}).$$

While cosine distance removes scale dependence, it still distinguishes vectors that differ by a sign, assigning maximal distance $\pi$ to antipodal points. This property becomes significant when representations are defined only up to sign or more general gauge transformations. Computationally, cosine angles are evaluated with clipping of inner products to $[-1,1]$ to avoid numerical artifacts due to floating-point errors.

**Projective geometry and the Fubini–Study metric.** We then define the projective distance used to identify gauge-equivalent representations. Projective geometry treats vectors that differ by a nonzero scalar as equivalent. In the real case, the relevant space is real projective space $\mathbb{RP}^{D-1}$, obtained by identifying $r \sim \lambda r$ for all $\lambda \neq 0$. A natural distance on this space is induced by the Fubini–Study metric. For normalized representatives $\hat{r}_k, \hat{r}_{k+1}$, the real Fubini–Study distance is defined as

$$d_{FS}(r_k, r_{k+1}) = \arccos\left(|\langle \hat{r}_k, \hat{r}_{k+1} \rangle|\right),$$

which takes values in $\left[0, \frac{\pi}{2}\right]$. The absolute value enforces the identification of antipodal points, so that vectors differing only by sign have zero distance. As before, cumulative projective drift is defined by

$$\Delta_{FS}(n) = \sum_{k=1}^{n-1} d_{FS}(r_k, r_{k+1}).$$

This construction measures change between equivalence classes rather than between raw vectors. In contrast to cosine distance, the Fubini–Study metric is invariant under both rescaling and sign inversion. Numerical implementation follows the same normalization and clipping procedures as for cosine distance, with the additional absolute value applied to inner products prior to evaluating the inverse cosine.

**Principal component analysis and induced projective ambiguity.** To ensure the presence of nontrivial projective equivalence, representations $r_k$ are derived as principal component directions from each data window. For a window $X_k \in \mathbb{R}^{W \times D}$, the data are centered by subtracting the mean vector $\mu_k = \frac{1}{W}\sum_{i=1}^{W} x_i$. Singular value decomposition is then applied,

$$X_k - \mathbf{1}\mu_k^\top = U_k \Sigma_k V_k^\top,$$

where $V_k \in \mathbb{R}^{D \times D}$ contains orthonormal right-singular vectors. The first principal component direction is $r_k = v_{k,1}$, the first column of $V_k$. By construction, $v_{k,1}$ is defined only up to a sign, since both $v_{k,1}$ and $-v_{k,1}$ correspond to the same variance-maximizing direction. This sign indeterminacy introduces genuine projective ambiguity into the representation trajectory. No sign alignment is performed across windows and the raw output of the decomposition is retained. This choice allows the comparison between cosine and Fubini–Study distances to directly reflect the treatment of antipodal equivalence.

**Detection of sign flips and gauge-induced drift along the trajectory..** Given successive principal component vectors $r_k$ and $r_{k+1}$, the inner product $\langle r_k, r_{k+1} \rangle$ is monitored. Negative values indicate a sign inversion between consecutive windows. These events are not corrected but are recorded as part of the trajectory. Under cosine distance, each sign flip contributes a jump of approximately $\pi$ to the cumulative drift. Under the Fubini–Study metric, the same event contributes zero, reflecting gauge equivalence. The cumulative difference

$$\Delta_C(n) - \Delta_{FS}(n)$$

therefore provides a quantitative measure of drift attributable solely to sign ambiguity. This difference is nonnegative and nondecreasing along the trajectory. Its computation requires no additional parameters beyond those already defined for the cosine and Fubini–Study distances.

**Logarithmic comparison between euclidean and projective drift.** To contrast raw magnitude-based drift with projective drift, the ratio between Euclidean and Fubini–Study cumulative measures is considered. For numerical stability, a small constant $\varepsilon > 0$ is added to both quantities and the logarithmic ratio is defined as

$$R(n) = \log_{10}\left(\frac{\Delta_E(n) + \varepsilon}{\Delta_{FS}(n) + \varepsilon}\right).$$

This quantity highlights orders-of-magnitude differences between Euclidean and projective change and is plotted as a function of the trajectory index. The logarithmic transformation facilitates visualization when Euclidean drift grows much faster than projective drift. The choice of ε is fixed and does not affect qualitative trends.

**Computational tools and implementation**. All computations are performed using standard scientific computing libraries. Linear algebra operations, including norms, inner products and singular value decompositions, are implemented using optimized numerical routines. Principal component extraction relies on singular value decomposition to ensure deterministic behavior. Vector normalization, cumulative summation and trigonometric evaluations are carried out in double precision. Visualization of trajectories and cumulative measures uses standard plotting libraries with explicit control over axes and scaling. No stochastic optimization, learning procedure or parameter tuning is involved beyond the selection of window size and step length, which are fixed prior to analysis.

In conclusion, we specified here the data constructions, geometric distances and computational steps to analyze representation drift under Euclidean, cosine and Fubini–Study metrics. By detailing explicit mathematical definitions and reproducible procedures, it allows comparison of raw, angular and projective notions of change in high-dimensional representation trajectories.

RESULTS

We report empirical observations obtained by applying Euclidean, cosine, and Fubini–Study distances to representation trajectories extracted from real high-dimensional data. We focus on quantitative differences between these measures in the presence of genuine projective ambiguity and on how cumulative drift separates into intrinsic and gauge-induced components along the same empirical paths.

**Empirical separation of cosine and projective drift**. Analysis of representation trajectories derived from sliding-window principal component directions reveals a systematic divergence between cosine and Fubini–Study cumulative drift when sign ambiguity is active (Fig. A–B). Across the full trajectory of 32 windows, the cumulative cosine drift reached 51.15 rad, whereas the cumulative Fubini–Study drift reached 14.72 rad, yielding a net difference of 36.43 rad attributable solely to sign inversions. Inspection of stepwise inner products between consecutive principal directions identified 17 sign-flip events, each corresponding to a negative dot product (Fig. D). Under cosine distance, each event contributes an increment close to $\pi$, while under the Fubini–Study metric the same event contributes zero. Outside these sign-flip events, incremental angular changes measured by cosine and Fubini–Study distances coincide numerically. A paired comparison of stepwise angular increments therefore reduces to 17 nonzero differences, all positive (exact two-sided sign test: $p<0.001$). These results show that, along a fixed empirical trajectory, cosine distance aggregates both intrinsic directional change and gauge-induced variability, whereas the Fubini–Study distance aggregates only the former. Therefore, the cumulative difference between the two acts as a statistically discriminable measure of gauge-induced drift, computed without alignment or post hoc correction.

**Relation between raw magnitude drift and projective change**. Comparison with Euclidean cumulative drift further clarifies the structure of representation change (Fig. A–C). The cumulative Euclidean drift between successive principal directions increased monotonically and substantially over the trajectory, reflecting sensitivity to both magnitude and direction. When contrasted with Fubini–Study drift, the logarithmic ratio $\log_{10}[(\Delta_E + \varepsilon)/(\Delta_{FS} + \varepsilon)]$ increased steadily along the trajectory (Fig. C), indicating that raw magnitude-based change grows at a substantially faster rate than intrinsic projective change. This divergence persists even in intervals where no sign flips occur, showing that Euclidean distance conflates smooth directional variability with scale and parametrization effects that are removed by normalization and projective identification. The monotonic increase of the logarithmic ratio confirms that the separation reflects accumulated structure rather than isolated events. Together with the statistically significant cosine–Fubini–Study separation, our analysis decomposes observed drift into three components measured on the same data: raw magnitude change, angular change sensitive to sign and intrinsic projective change insensitive to gauge.

Overall, our results show that cosine and Fubini–Study distances coincide except at empirically detected sign-flip events, where cosine drift accumulates large increments that are absent under the Fubini–Study metric. These increments are statistically significant and account for a substantial fraction of total angular drift. In parallel, Euclidean drift grows much faster than projective drift, producing a steadily increasing logarithmic separation between raw and intrinsic measures of change.

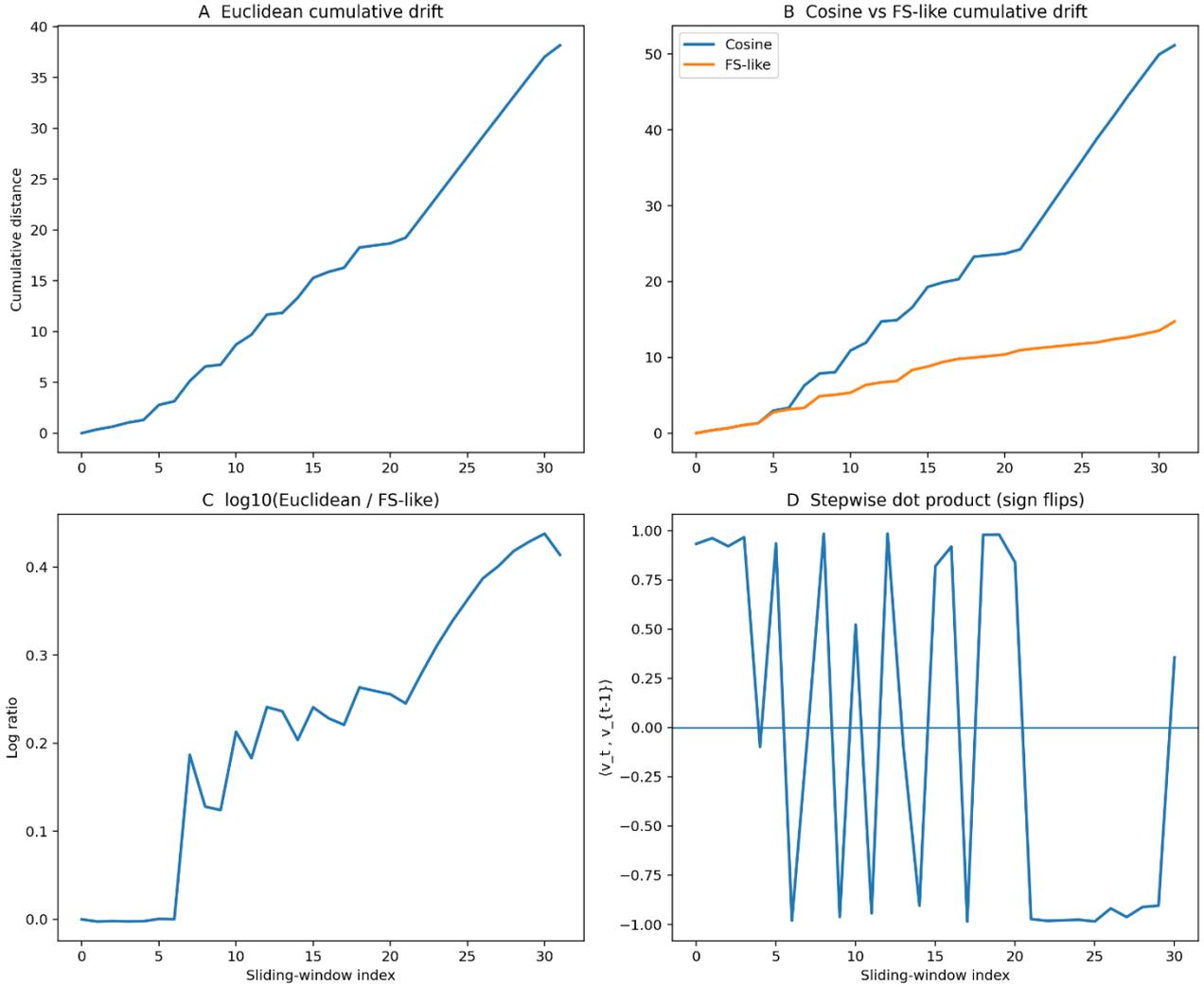

**Figure**. Projective effects in empirical high-dimensional data.
(A) Cumulative Euclidean drift between successive first principal components (PC1) computed on sliding windows of handwritten-digit images. Euclidean distance grows steadily, reflecting sensitivity to parametrization changes.
(B) Cumulative angular drift comparing cosine distance and a projective (FS-like) distance. Cosine drift accumulates large jumps due to repeated sign inversions of PCA eigenvectors, whereas the FS-like metric remains insensitive to these gauge transformations.
(C) Logarithmic ratio between Euclidean and FS-like cumulative drift, highlighting an increasing separation between raw distance growth and intrinsic projective change.
(D) Stepwise dot product between consecutive PC1 vectors; negative values indicate eigenvector sign flips ($v \equiv -v$), the source of genuine projective ambiguity.
Together, the panels show that Euclidean and cosine metrics overestimate change when representations are defined only up to sign, while the FS-like metric isolates intrinsic evolution of equivalence classes.

CONCLUSIONS

We found that representation drift in high-dimensional data decomposes into distinct geometric components that are conflated by commonly used distance measures. By examining the same empirical trajectories with Euclidean, cosine and Fubini–Study distances, we showed that raw magnitude-based drift grows rapidly and monotonically, angular drift aggregates both intrinsic directional change and gauge-induced variability and projective drift isolates only the intrinsic component. Quantitatively, the divergence between cosine and Fubini–Study cumulative drift was entirely attributable to empirically observed sign inversions of principal component directions, while the remaining increments coincided exactly. The resulting difference constituted a substantial fraction of total angular drift and was statistically discriminable under an exact test, demonstrating that the separation is not a numerical artifact. In parallel, the steadily increasing logarithmic gap between Euclidean and projective drift indicated that raw distances systematically overestimate change even when no discrete gauge events occur. Together, these observations provide an empirical picture in which different metrics respond to different aspects of representation variability and in which projective geometry provides a way to factor out equivalence-induced fluctuations while retaining sensitivity to genuine evolution. This synthesis of quantitative results establishes a clear basis for interpreting representation drift as a composite phenomenon across heterogeneous representations.

We reframe representation drift as a geometric problem defined on equivalence classes rather than on fixed coordinate vectors. Existing techniques address invariances through preprocessing, normalization or alignment steps applied before distance computation, assuming that a single corrected representation suffices. In contrast, the use of the Fubini–Study metric incorporates invariance directly into the distance itself, ensuring that gauge-equivalent representations are identified by construction. This distinction clarifies why cosine distance, while scale invariant, remains sensitive to sign and related ambiguities and why Euclidean distance is sensitive to both scale and parametrization. We do not introduction a new similarity score but rather provide the explicit separation of intrinsic and gauge-induced contributions to drift along the same empirical trajectory. Compared with information-theoretic or distribution-based measures, our projective metric operates directly on representations and does not require probabilistic modeling (Vasudevan et al. 2018; Sechidis et al. 2018; Liu et al. 2022). Compared with alignment-based methods, it avoids reference choices and post hoc corrections (Galpert et al. 2018; Kobus et al. 2020; Expósito et al. 2022; He et al. 2024). Our approach is neither a replacement for standard distances nor an ad hoc correction, but a geometric refinement becoming active when representations are defined up to equivalence. In this sense, our method can be classified as a projective geometric approach to representation analysis, complementary to Euclidean, angular and statistical methods and naturally situated between raw vector-space techniques and more abstract manifold-based analyses.

Several limitations must be acknowledged. First, our empirical demonstration relies on a dataset that is not intrinsically temporal; trajectories are constructed through sliding windows and the observed drift reflects this imposed ordering. While this procedure is mathematically well defined, it does not capture all sources of variability present in naturally evolving systems. Second, projective ambiguity is introduced through principal component analysis and the frequency of sign inversions depends on window size and overlap. Third, statistical testing is limited to exact tests on detected sign-flip events within a single trajectory and does not establish population-level generality. Fourth, the analysis focuses on real projective geometry associated with sign equivalence; more general gauge freedoms, such as rotations in latent spaces, are not explored empirically here. Finally, the figures illustrate a representative case rather than a systematic survey across datasets or representations.
Our results point toward several concrete directions for application and further investigation. The projective decomposition of drift can be applied to representations known to exhibit gauge freedom, such as eigenmodes of covariance or connectivity matrices, latent variables in dimensionality reduction or embeddings obtained from independent training runs. Future studies could test whether the cumulative difference between cosine and Fubini–Study drift systematically increases with the degree of representation non-identifiability, providing a quantitative signature of gauge freedom. Another testable hypothesis is that intrinsic projective drift stabilizes earlier than raw or angular drift along learning or preprocessing pipelines, indicating convergence of representation content despite ongoing parametrization changes. Methodologically, this calls for reporting projective drift alongside conventional distances when representations are not uniquely defined and for explicitly testing whether observed changes persist under projective identification.

In summary, we addressed the problem of distinguishing intrinsic representation change from variability arising from non-identifiability in high-dimensional data. By comparing Euclidean, cosine and Fubini–Study distances along the same empirical trajectories, we showed that projective geometry could provide a computable way to make this distinction explicit. Representation drift is not a single quantity, but a composite of raw, angular and projective components, each capturing a different aspect of change. Recognizing and separating these components could allow a more precise interpretation of how representations evolve, providing a geometric criterion that is mathematically grounded and empirically testable.

**Competing interests.** The Author does not have any known or potential conflict of interest including any financial, personal or other relationships with other people or organizations within three years of beginning the submitted work that could inappropriately influence or be perceived to influence their work.
**Funding.** This research did not receive any specific grant from funding agencies in the public, commercial or not-for-profit sectors.
**Acknowledgements:** none.
**Authors' contributions.** The Author performed: study concept and design, acquisition of data, analysis and interpretation of data, drafting of the manuscript, critical revision of the manuscript for important intellectual content, statistical analysis, obtained funding, administrative, technical and material support, study supervision.
**Declaration of generative AI and AI-assisted technologies in the writing process.** During the preparation of this work, the author used ChatGPT 5.2 to assist with data analysis and manuscript drafting and to improve spelling, grammar and general editing. After using this tool, the author reviewed and edited the content as needed, taking full responsibility for the content of the publication.